\documentclass[journal]{IEEEtran}
\usepackage{amsmath}
\usepackage{lineno,hyperref}
\usepackage{booktabs}
\usepackage{multirow}
\usepackage{graphicx}
\usepackage[justification=centering]{caption}
\ifCLASSINFOpdf
\else
\fi
\begin{document}
%
\title{Downhole Track Detection via Multiscale Conditional Generative Adversarial Nets}

\author{Jia Li,~\IEEEmembership{Student Member,~IEEE,}
       Xing Wei,~\IEEEmembership{Member,~IEEE,}
        Guoqiang Yang,~\IEEEmembership{Student Member,~IEEE,}
        Xiao Sun,~\IEEEmembership{Member,~IEEE,}
        Changliang Li,~\IEEEmembership{Member,~IEEE, and}
\thanks{J. Li, X. Wei, X. Sun and Guoqiang Yang are with the school of Computer and Information, Hefei University of Technology,
 Hefei 230009, China. E-mail: lijiajia@mail.hfut.edu.cn; weixing@hfut.edu.cn ygqhfut@163.com; sunx@hfut.edu.cn}
\thanks{C. Li is currently the head with the Kingsoft institute of Artificial Intelligence. E-mail: lichangliang@kingsoft.com.}
}
\maketitle
\begin{abstract}
Frequent mine disasters cause a large number of casualties and property losses. Autonomous driving is a fundamental measure for solving this problem, and track detection is one of the key technologies for computer vision to achieve downhole automatic driving. The track detection result based on the traditional convolutional neural network (CNN) algorithm lacks the detailed and unique description of the object and relies too much on visual postprocessing technology. Therefore, this paper proposes a track detection algorithm based on a multiscale conditional generative adversarial network (CGAN). The generator is decomposed into global and local parts using a multigranularity structure in the generator network. A multiscale shared convolution structure is adopted in the discriminator network to further supervise training the generator. Finally, the Monte Carlo search technique is introduced to search the intermediate state of the generator, and the result is sent to the discriminator for comparison. Compared with the existing work, our model achieved 82.43\% pixel accuracy and an average intersection-over-union (IOU) of 0.6218, and the detection of the track reached 95.01\% accuracy in the downhole roadway scene test set.
\end{abstract}


\ifCLASSOPTIONpeerreview
\begin{center} \bfseries EDICS Category: 3-BBND \end{center}
\fi
\IEEEpeerreviewmaketitle
\section{Introduction}

\IEEEPARstart{I}{n} recent years, the frequent occurrence of large-scale mine accidents has caused a large number of casualties and property losses. The production and transportation in mining need to be developed in an unmanned and intelligent direction.
Track detection is one of the key technologies in computer vision for underground automatic driving. Track detection refers to recognizing the track area in a video or image by image processing technology, which shows the specific position of the track line. It can assist in the detection of pedestrians and obstacles and further improve the driving safety of underground locomotives. However, underground track detection is easily affected by complex environmental factors, such as light changes, water cover and cable interference; thus, in recent years, track detection has become a challenging task in studying computer vision.

Track detection algorithms based on traditional image processing can be roughly divided into two categories: feature-based methods and model-based methods. Feature-based track detection technology \cite{[1],[2]} mainly uses feature information such as track edge, texture, color, geometry and gray value to distinguish the track area from the surrounding environment. However, this method relies too much on the underlying features of the image and the surrounding environment easily interferes. The basic principle of the model-based track detection method \cite{[3]} is to transform the track detection problem into a problem of solving the track model parameters. According to the track pattern in the local area, the fitting of the track line is achieved by using a segmentation line to describe the model. However, the algorithm lacks the robustness and flexibility for any road shape.

Recently, the deep convolution neural network (DCNN) has been successfully applied to many different computer vision tasks. The problem of track line detection is solved as an image segmentation task \cite{[4],[5]}. The network predicts the pixels at the position of the track, then combines the pixels of the same track, and finally displays the position of the track line in the target image. However, the problem of downhole track detection scenes is not a direct classification task for track line pixels. Moreover, the prediction of the track line needs to preserve the structure or quality of the equivalent track, that is, the fineness and uniqueness of the track line. In addition, in the process of training, it is necessary to manually design a complex loss function that is suitable for improving the final detection effect.

Another method for solving the above problem is to use a generative adversarial network (GAN) \cite{[6]}. The GAN contains two opposing models: a generative model G for fitting the sample data distribution and a discriminative model D for judging the true and false data. However, one of the disadvantages of GANs is that data generated are uncontrollable.
The conditional generative adversarial network (CGAN) \cite{[7]} adds an additional conditional y to generator and discriminator on the basis of a GAN. The generator must generate a sample that matches the condition y. The discriminator must determine not only whether the image is true but also whether the image and the condition match. Some scholars have achieved good results in image generation via CGANs. Isola et al. \cite{[9]} proposed a network called the pix2pix framework for paired image transformation based on a CGAN. This method has achieved good results. However, the model is limited to generating low-resolution images. Chen and Koltun \cite{[10]} used modified perceptual loss \cite{[11],[12],[13]} to generate images. Although models can generate high-resolution images, generated images often lack fine detail and realistic texture.

In view of the shortcomings of previous works, this paper proposes a downhole track line detection model based on a CGAN. We use the method of adversarial learning to solve the problem of artificially designing complex loss functions and introduce a Monte Carlo search to solve the problem of generating image distortion and lack of precision. In summary, the paper makes the following contributions:

\begin{itemize}{
\item This paper introduces the method of adversarial learning in the field of track line detection for the first time and proposes generator and discriminator networks based on a multiscale structure so that the generator obtains sufficient global and local information in the process of generation.
\item To strengthen the constraints in the image generation process, this paper introduces the Monte Carlo search technology to the generator and solves the problem of image distortion and lack of detail.
\item To promote the fusion of global and local information, this paper introduces a multitask learning strategy based on parameter sharing in the discriminator network, which indirectly expands the storage capacity of the discriminator model and accelerates the model convergence.
\item The experiments prove that compared with the previous work, the accuracy of the track recognition of the proposed model reaches 95.01\%, an image with resolution up to 2K can be generated, and the details and texture of the images are greatly improved.
}\end{itemize}

The structure of the rest of the paper is as follows. The second section introduces the proposed model. The third section shows the experiment preparation. The fourth section analyzes the experimental results, and the fifth section gives the conclusions and future work.

\section{Related Work}
\subsection{Image-to-image translation}

Many researchers have leveraged adversarial learning for image-to-image translation \cite{[9]}, which translates an input image from one domain to another domain given input-output image pairs as training data. CGANs aim to model the conditional distribution of real images given the input semantic label maps via the following minimax game:
\[\mathop {\min }\limits_G \mathop {\max }\limits_D {{\cal L}_{GAN}}(G,D)\]
where the objective function $L_{GAN}(G,D)$ is given by:
\[{E_{s,x}}[\log D(s,x)] + {E_s}[\log (1 - D(s,G(s)))]\]
Adversarial loss has become a popular choice for many image translation tasks because the discriminator can learn the trainable loss function and automatically adapt to the differences between the generated and real images in the target domain.
For the purpose of this paper, the goal is to input an image containing a downhole track line, and generator G generates an image that marks the existing track line. In other words, the training dataset is given as a set of pairs of corresponding images \({(s_i,x_i)}\),where \(s_i\) is a semantic label map, and \(x_i\) is the corresponding natural image. However, in the application of downhole roadway scenes, the results generated by pix2pix are limited to low-resolution images. The generated image still has a large gap from the real sample, and the image still lacks texture and details.

\subsection{Track line detection}

We also discuss other related methods for track detection. In the feature-based approach, Quach et al.  \cite{[1]} proposed color and depth information recorded using a single RGB-D camera to better handle unfavorable factors such as lighting conditions. However, since this method relies on the underlying features of the image, environmental factors easily interfere, making the algorithm less robust. Model-based methods, such as Bente et al. \cite{[3]} proposed a lane detection method using the Hough transform and contour detection. They determine the corresponding model parameters by analyzing the target information in the road image. However, a road model often cannot adapt to multiple road conditions at the same time.

Additionally, some scholars have proposed using DCNNs to detect lane lines. In \cite{[4]}, Pan trained a spatial convolutional neural network (CNN) for specific problems and added postprocessing techniques that rely on handcrafting. Another recent example is the work of Neven et al. \cite{[5]}, which first used a segmentation network to obtain a lane marker prediction map. The second network was then trained to perform a constrained perspective transformation, and finally, the network used curve fitting to obtain the final result. However, their method relies on more postprocessing techniques, which increases the complexity of the actual application of the model.

\section{Multi-dimensional Generative Adversarial Model}

\subsection{Multi-granularity generator}

\begin{figure*}[htp]
	\begin{center}
		\includegraphics[width=0.95\textwidth]{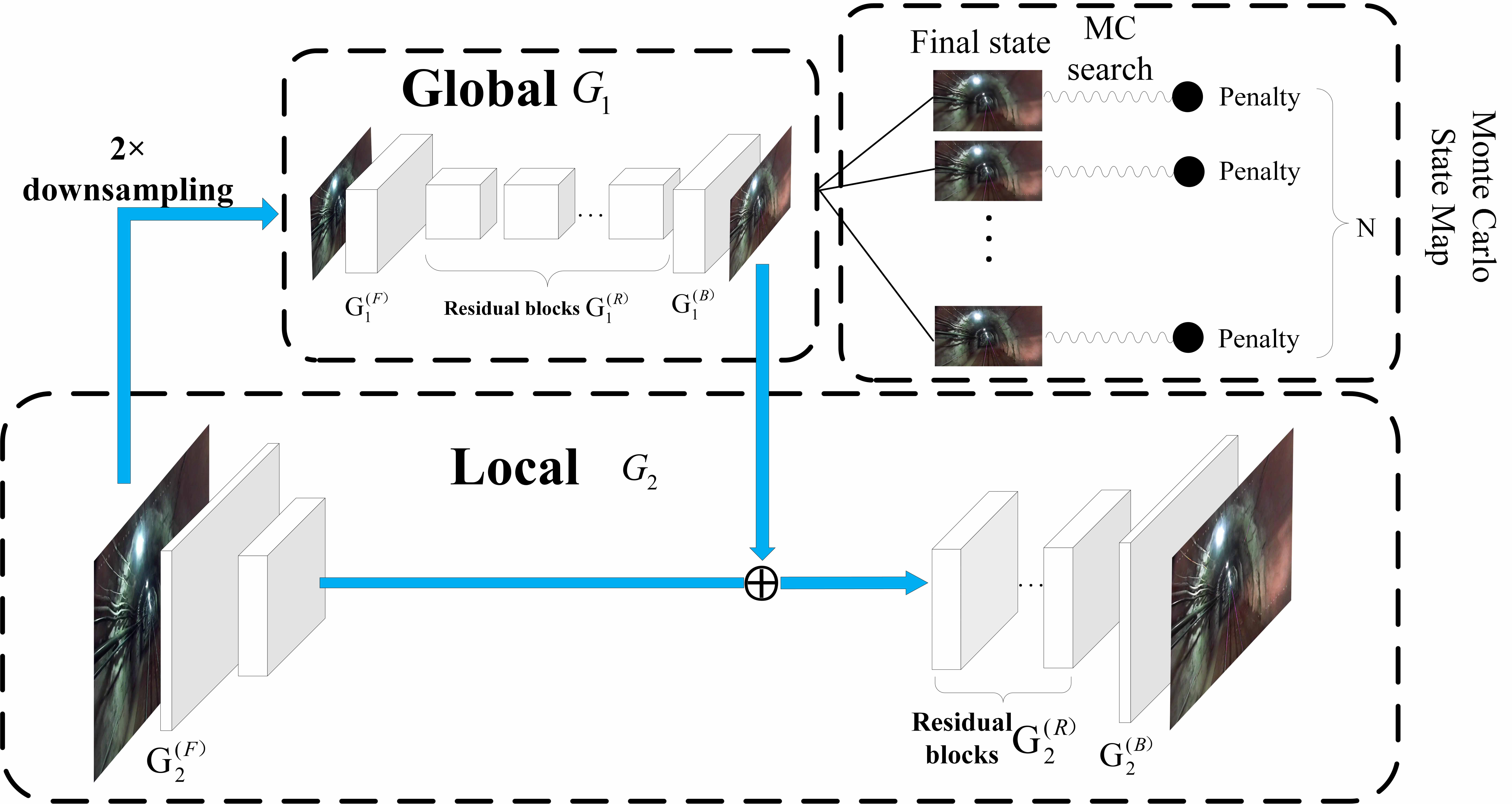}
		\caption{Multigranularity generator network structure}\label{fig:002}
	\end{center}
\end{figure*}

Referring to the hierarchical reinforcement learning in \cite{[20]}, we decompose the generator into two sub-generators \(G_1\) and \(G_2\), where \(G_1\) is the global generator, \(G_2\) is the local generator, and the overall structure of the generator \(G=\{G_1,G_2\}\) is shown in Fig. \ref{fig:002}. The global generator is mainly used for the overall information construction of images. The local generator can effectively improve the resolution of the generated image.

The global generator of the model proposed in this paper is designed based on the work in \cite{[13]}. It consists of 3 components: a convolutional frontend \({{G}}_1^{(F)}\), a set of residual blocks \({{G}}_1^{(R)}\), and a transposed convolutional backend \({{G}}_1^{(B)}\). A semantic label map of resolution 1024*512 is passed through the 3 components sequentially to output an image of resolution 1024*512. The local enhancer network also consists of 3 components: a convolutional frontend \({{G}}_2^{(F)}\), a set of residual blocks \({{G}}_2^{(R)}\), and a transposed convolutional backend \({{G}}_2^{(B)}\). The resolution of the input image to \(G_2\) is 2048*1024. Different from the global generator network, the input to the residual block \({{G}}_2^{(R)}\) is the elementwise sum of two feature maps: the output feature map of \({{G}}_2^{(F)}\) and the last feature map of the backend of the global generator network \({{G}}_1^{(B)}\), which helps to integrate the global information from \(G_1\) to \(G_2\).

\subsection{Multi-scale shared convolution discriminator}

\begin{figure*}[htp]
	\begin{center}
		\includegraphics[width=0.95\textwidth]{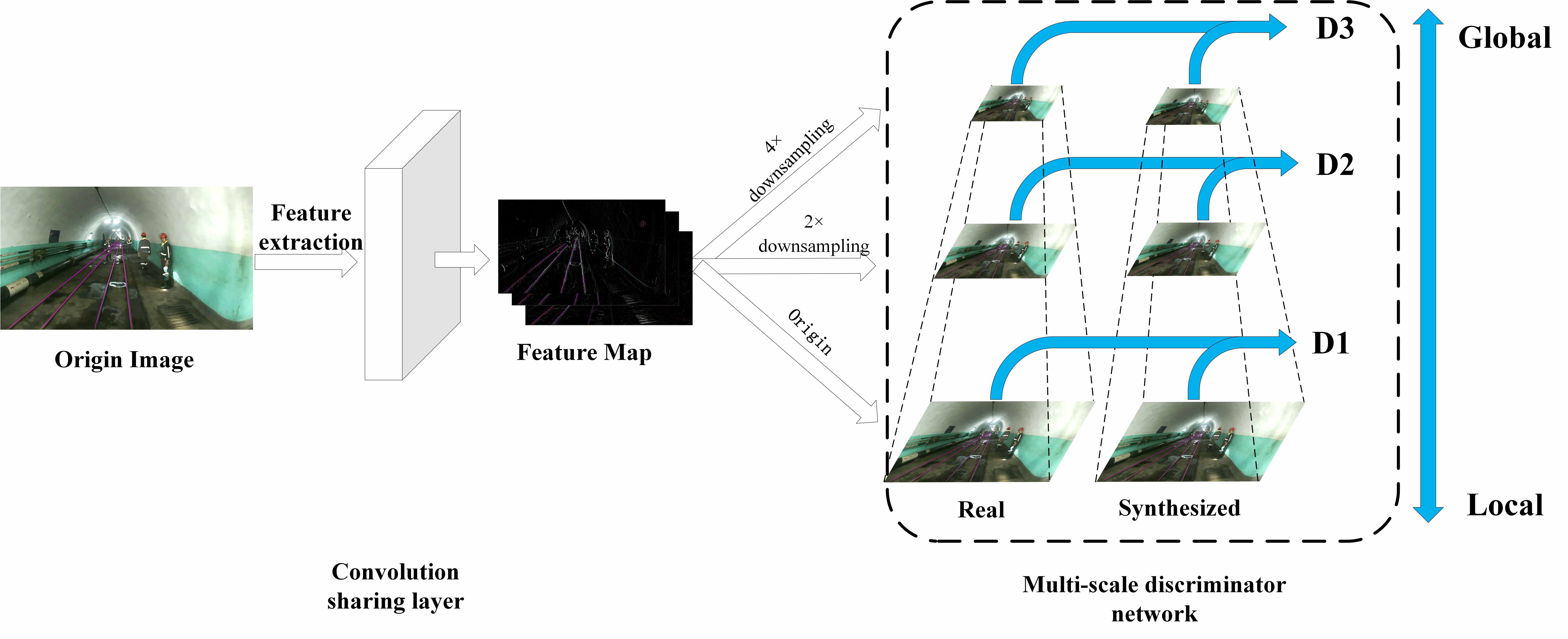}
		\caption{Multiscale shared convolution discriminator structure}\label{fig:003}
	\end{center}
\end{figure*}
To differentiate high-resolution real and synthesized images, we propose multiscale discriminators. We use 3 discriminators that have an identical network structure but operate at different image scales and the overall structure of the discriminators is shown in Fig. \ref{fig:003}. To promote the learning of each discriminator, the model introduces a multitask learning strategy based on parameter sharing \cite{[21]}. We first extract the images and generate the primary features of the images through a shared convolutional layer and obtain the corresponding feature map. Then, the feature samples of the real sample and the generated sample are downsampled using 2 and 4 as sampling factors, respectively. The three discriminators \(D_1\), \(D_2\), and \(D_3\), are also used to process these three different scale images. A discriminator with a large input scale has a more global perception of the image and can guide the global generation of the image. A discriminator with a smaller input scale is better at guiding the details of the generated image to further improve the overall image. In addition, the introduced multitask learning strategy greatly increases the storage capacity of the discriminator, so that the discriminator has more memory for learning how to discriminate the image and accelerate the convergence of the model. The specific process is as follows:
\[\mathop {\min }\limits_G \mathop {\max }\limits_{{D_1},{D_2},{D_3}} \sum\limits_{k = 1,2,3} {{{\cal L}_{{\rm{GAN}}}}(G,{D_k})} \]

where \(D_k\) represents one of the three discriminators.

\subsection{Optimization algorithm based on a Monte Carlo search}

We note that such a multiresolution pipeline is a well-established practice in computer vision \cite{[22],[23]}, and a two-scale pipeline is often enough. The experimental results show that although the performance improved, the generated images have the disadvantages of blurring and ghosting.
Thus, this paper introduces the Monte Carlo search technology to the model so that the generator can obtain guidance information rapidly when generating images. By searching the intermediate state of the generator multiple times and then sending the search results to the discriminator to calculate the penalty values, the generator constraints in the generation process are strengthened, and the quality of the generated image is further improved. The search process is shown in Fig. \ref{fig:002}.
First, the model uses G to perform a Monte Carlo search on the intermediate state of the generator. The specific process is as follows:
\[{\rm{\{ }}{Y^1},...,{Y^N}\}  = M{C^{{G_\beta }}}({Y_{1:t}};N)\]

where N represents the number of searches. $M C^{G_{\beta}}$ represents the state of the simulation using the Monte Carlo search. $G_{\beta}$ represents another generator virtualized by the Monte Carlo search technology, which has the same parameters as the actual generator. $Y_t$ represents the intermediate state to be sampled. $Y_{t+1}^i$ represents the final state obtained after sampling. After obtaining the N sampling results, the final states are sent to the discriminator, and the corresponding loss value $Q_D^G$ is calculated by equation (1). The specific process is as follows:

\[Q_D^G = \frac{1}{N}\sum\limits_{n = 1}^N {\sum\limits_{k = 1,2,3} {{\cal L}(Y_{t + 1}^i,{D_k})} } ,Y_{t + 1}^i \in M{C^{{G_\beta }}}({Y_t};N)\]
where \(D_k\) represents one of the three discriminators. To further strengthen the constraint, we refer to the perceptual loss in \cite{[11],[12],[13]} and propose a feature matching loss. The model extracts features from different layers of the discriminator and learns to match these intermediate states. Here, \(D_k(i)\) is defined to represent the i-th layer feature extractor of discriminator \(D_k\); then, the feature matching loss can be expressed as:
\[{{\cal L}_{FM}}(G,{D_k}) = {{\rm E}_{(s,x)}}\sum\limits_{i = 1}^T {\frac{1}{{{N_i}}}[\left\| {D_k^{(i)}(s,x) - D_k^{(i)}(s,G(s))} \right\|} ]\]
where T represents the number of network layers and \(N_i\) represents the number of elements per layer.
In summary, the final loss function is:
\[\mathop {\min }\limits_G (\mathop {\max }\limits_{{D_1},{D_2},{D_3}} Q_G^D + \lambda \sum\limits_{k = 1,2,3} {{{\cal L}_{FM}}(G,{D_k}))} \]
where \(\lambda\) represents the manually set weighting factor. It is worth noting that in feature matching loss, \(D_k\) is used as a feature extractor and does not maximize loss.

\section{Experiments Preparation}

\subsection{Datasets}
Since there are currently no public datasets containing downhole track lines, we use video cameras fixed on mine locomotives to collect video data. The datasets include different scenarios from multiple mines.
In actual data processing, we use data enhancement technology to expand the datasets. The specific transformations include image rotation transformation, mirror transformation, flip image transformation and other methods to effectively expand the datasets. We finally obtained approximately 2,500 pictures. The training set and the validation set are then divided in an 8:2 manner. More data can be obtained at https://github.com/LJ2lijia/Downhole-track-line-dataset.
\subsection{Metrics}
The experiments use the official indicators on the ground \cite{[24]}, namely, \(Acc\) (accuracy), \(FP\) (false positive), and \(FN\) (false negative), which are defined as follows:
\[Acc = \sum\limits_{im} {\frac{{{C_{im}}}}{{{S_{im}}}}} \]

where \(C_{im}\) is the number of correct prediction points generated during the test, and \(S_{im}\) is the number of ground truths. When the distance between the predicted point and the real point is less than the set threshold (here, the threshold is set to 3), the point is considered correct.
\[FP = \frac{{{F_{pred}}}}{{{N_{pred}}}}\]
\[FN = \frac{{{M_{pred}}}}{{{N_{gt}}}}\]
where \(F_{pred}\) is the erroneously predicted track line, \(N_{pred}\) is the track line that needs to be predicted, \(M_{pred}\) is the track line that is mistaken for the ground truth, and \(N_{gt}\) is the number of all track lines.

\subsection{Training Details}
All the networks are trained from scratch using the Adam solver and a learning rate of 0.05. We keep the same learning rate for the first 100 epochs and linearly decay the rate to zero over the next 100 epochs. Weights are initialized from a Gaussian distribution with a mean of 0 and a standard deviation of 0.02. The number of Monte Carlo searches N is set to 5, and the specific gravity $\lambda$ between the control feature matching loss and the discriminator loss function is set to 10. The implementation of our model is based on the PyTorch deep learning framework.

\section{Experiments}

\subsection{Automatic Evaluation}

To compare the model proposed in this paper with the existing ground lane detection algorithm, we transplant the existing ground lane detection algorithm to the underground for detecting the track line. The models involved in the contrast experiment are pix2pix in \cite{[9]}, spatial CNN (SCNN) in \cite{[4]}, LaneNet in \cite{[5]}, and segmentally switchable curves (SSC) in \cite{[26]}. The average value of the multiple experiments is shown in Table 2.

\begin{table}[htp]
	\label{tab:table_2}
	\centering
	\caption{Scores for testing our datasets using different methods}
	\begin{tabular}{cccc}
		\toprule
		Method        & Accuracy(\%)  & FP       & FN    \\
		\midrule
		SCNN	         & 93.26         & 0.0598  & 0.0269\\
		LaneNet      	 & 92.87         & 0.0620   & 0.0312\\
		SSC	         & 89.64	     & 0.0643   & 0.0393\\
		pix2pix	     & 92.89         & 0.0535   & 0.0297\\
		Ours	         & \bf{95.01}    & \bf{0.0401}	& \bf{0.0186}\\
		
		\bottomrule
	\end{tabular}
\end{table}

As shown in Table \uppercase\expandafter{\romannumeral1}, the model proposed in this paper obviously exceeds the previously existing models for this type of problem, both in terms of accuracy, FP and FN, which proves the superiority of the algorithm in this paper.

To verify the validity of the Monte Carlo search, we compare the model proposed in this paper with the model with the Monte Carlo search removed and explore the impact of the number of searches on the Monte Carlo search on the performance of the model where without MC represents the model with the Monte Carlo search removed, and N represents the number of Monte Carlo searches.

\begin{table}[htp]
	\label{tab:table_3}
	\centering
	\caption{Scores for testing our datasets using different methods}
	\begin{tabular}{ccccc}
		\toprule
		Method        & Accuracy(\%)   & FP       & FN     & Average Time(s)  \\
		\midrule
		Without MC  &	90.25	& 0.1031&	0.1002&	0.2567 \\
		N=1	& 92.87  &	0.0901  & 0.0912&	0.2678 \\
		N=3	& 93.09  &	0.0765  & 0.0703&	0.2806 \\
		N=5	& 95.01  &	0.0401  & 0.0186&	0.2962 \\
		N=7	& 95.68  &	0.0399  & 0.0176&	0.3465 \\
		N=9	& 95.96  &	0.0365  & 0.0170&	0.4031 \\
		
		\bottomrule
	\end{tabular}
\end{table}

As shown in Table \uppercase\expandafter{\romannumeral2}, it can be seen that the introduction of the Monte Carlo search obviously significantly improves the accuracy, FP and FN of the final generated results. We found that with the increase in the number of searches, when N=5, both time-consumption and accuracy achieve better results. As N continues to increase, the accuracy, FP, and FN improve, but the increase is not large, and the average generation time of each image increased considerably because as the number of searches increases, the number of calculations required to generate each image gradually increases, and the increase is nonlinear, so the number of Monte Carlo searches is set to 5 by default.

To verify the effect of multiscale discriminator and multitask learning, we conduct a comparative experiment on the model proposed in this paper, the model with a multiscale discriminator network but no multitask learning and the network using only a single-scale discriminator. The generator and loss functions are fixed during this experiment. The experimental results are shown in Table 3. To avoid the contingency of the experiment, we repeat the experiment several times, and the data in the table are the average of the experimental data.

\begin{table}[htp]
	\label{tab:table_3}
	\centering
	\caption{Scores for testing our datasets using different methods}
	\begin{tabular}{cccc}
		\toprule
		& Single D  & Multiscale Ds       & Ours     \\
		\midrule
		Pixel accuracy & 80.08	& 81.43 &\bf{82.68}  \\
		IOU	&  0.5125	&0.5818	&\bf{0.6351} \\
		\bottomrule
	\end{tabular}
\end{table}

From Table \uppercase\expandafter{\romannumeral3}
, we find that multiscale discriminator and multitask learning strategy can significantly improve the accuracy of the generated pixels because the multiscale discriminator makes the judgment of the generated pixel points stricter during the network training. The multitasking learning strategy allows the discriminator to learn more features, giving the generator more accurate guidance.

\subsection{Manual Evaluation}

To further evaluate the model proposed in this paper, we adopt the method of manual evaluation. The existing platform for manual evaluation is MTurk (Amazon Mechanical Turk), so we use a similar method to publish the results of the experiment online to a website and send them to volunteers through social platforms.

\begin{figure}[htp]
	\begin{center}
		\includegraphics[width=0.45\textwidth]{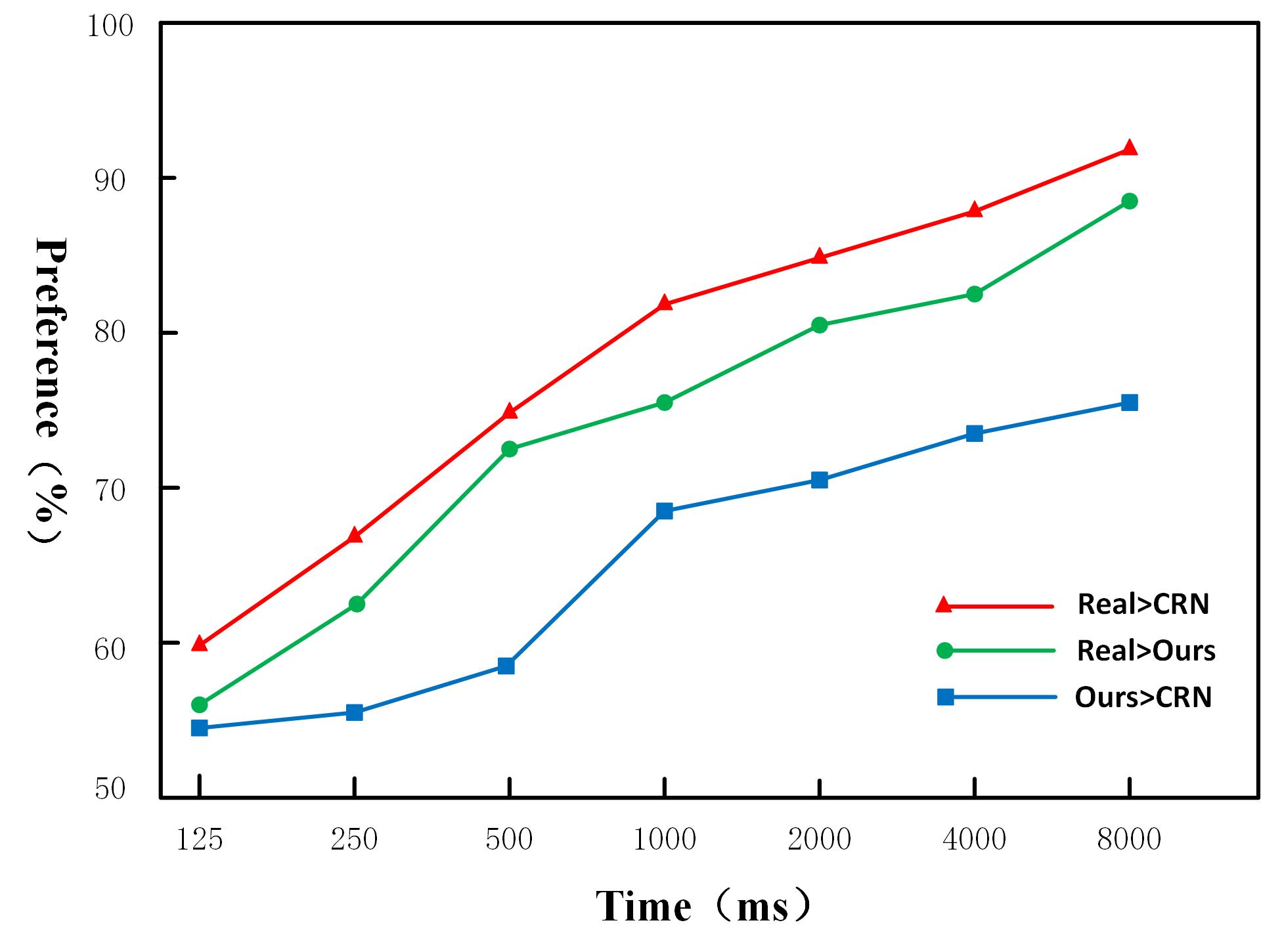}
		\caption{Preference-time fluctuation graph}\label{fig:005}
	\end{center}
\end{figure}

For this task, based on the same input image, we use the CRN model and the model proposed in this paper to generate two images and participate in the comparison with the real image. To better reflect fairness, we sent two of the three images to volunteers almost simultaneously. The volunteers are required to select the most accurate and texture-clear images within a limited time. The limited time is from 125 ms to 8000 ms. The comparison results are shown in Fig. \ref{fig:005}.
\begin{figure*}[htp]
	\begin{center}
		\includegraphics[width=0.9\textwidth]{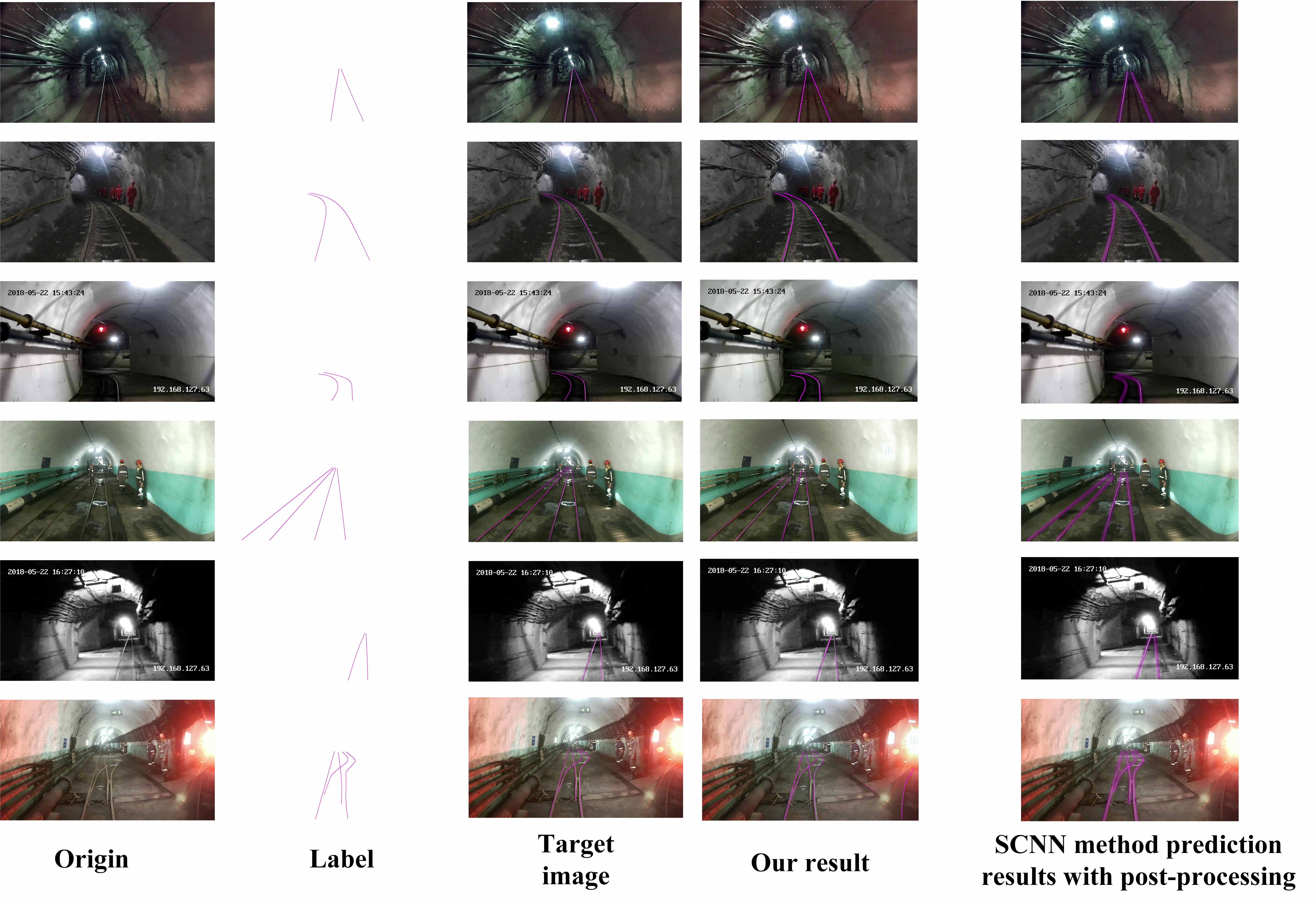}
		\caption{Comparison of our model and SCNN network test results}\label{fig:006}
	\end{center}
\end{figure*}

It can be seen from Fig. \ref{fig:005} that as the limited time increases, the difference between the three images becomes more apparent. The final result shows that the model in this paper is obviously better than the CRN model, and the gap with the real images becomes increasingly smaller.
\begin{figure*}[htp]
	\begin{center}
		\includegraphics[width=0.8\textwidth]{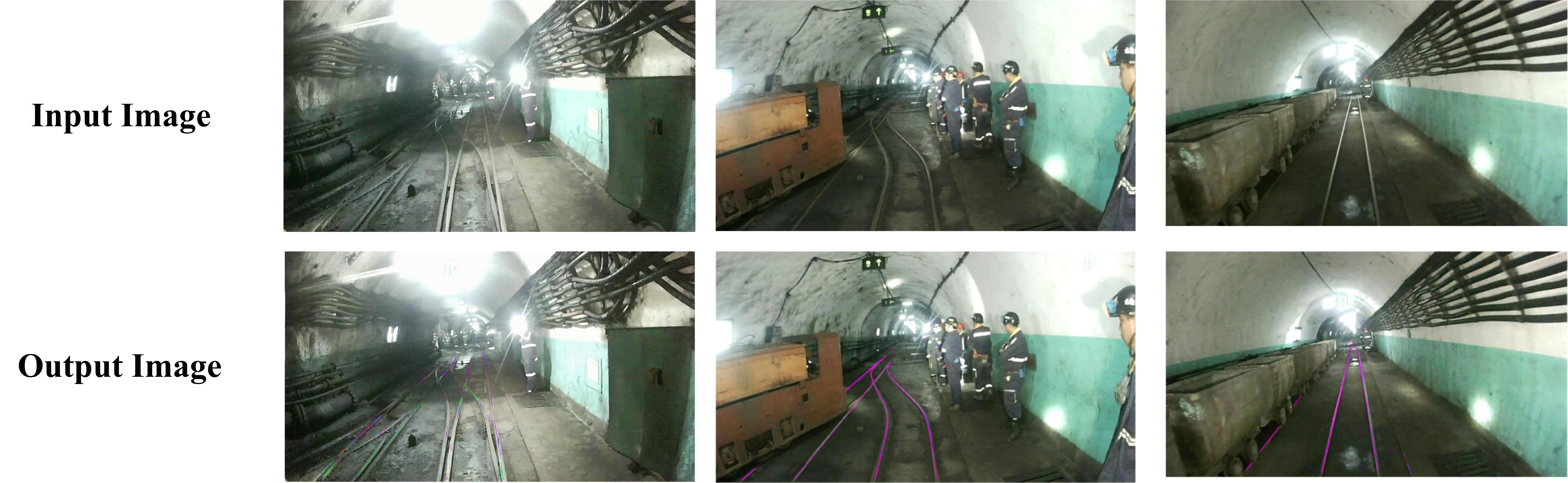}
		\caption{The test result of our model in the complex scenes}\label{fig:007}
	\end{center}
\end{figure*}
\section{Conclusion and Future Work}

\subsection{Case Study and Error Analysis}

We conducted a comparative analysis experiment with the SCNN in the downhole roadway scenario. Fig. \ref{fig:006} shows some of the test cases. It can be seen that the images generated by our model are very detailed, and high-quality images can be generated for all the above scenarios.  Our model can still accurately detect the tracks that are not marked in the training images, which fully demonstrates that the model has excellent robustness. The application of SCNN to the downhole scene also obtains good recognition results, but the robustness of the model is poor. In addition, the SCNN algorithm requires more postprocessing techniques, which improves the complexity of the visualization operation, and the detected results are not real. It can be seen from the above comparison experiments that our model has great advantages.

However, in the experiment, it is also found that if the scene in the image is very complicated (for example, more than four track lines and track lines are occluded), the result of our model will be affected.  Because the track line recognition in complex scenes requires more adequate and accurate guidance information, but the repetitive search method for intermediate states does not effectively constrain the generation of generators for some complex scenes. The experimental results are shown in Fig. \ref{fig:007}.

This paper proposes a downhole track line detection model based on a conditional adversarial generation network. The model realizes the generation of high-resolution downhole track line detection images by decomposing the generator into global and local generators. Then, the multiscale shared convolution structure is used in the discriminator network to further improve the overall image. The model uses Monte Carlo search technology to further improves the quality of the generated images. The experiments show that compared with traditional neural network detection models, the proposed model has better performance in terms of accuracy and other aspects and does not depend on cumbersome postprocessing techniques.
In the future, we will focus on the detection of track lines in complex scenes and further improve the speed and robustness of the model.

%
%
%

\ifCLASSOPTIONcaptionsoff
  \newpage
\fi

\begin{IEEEbiography}[{\includegraphics[width=1in,height=1.25in,clip,keepaspectratio]{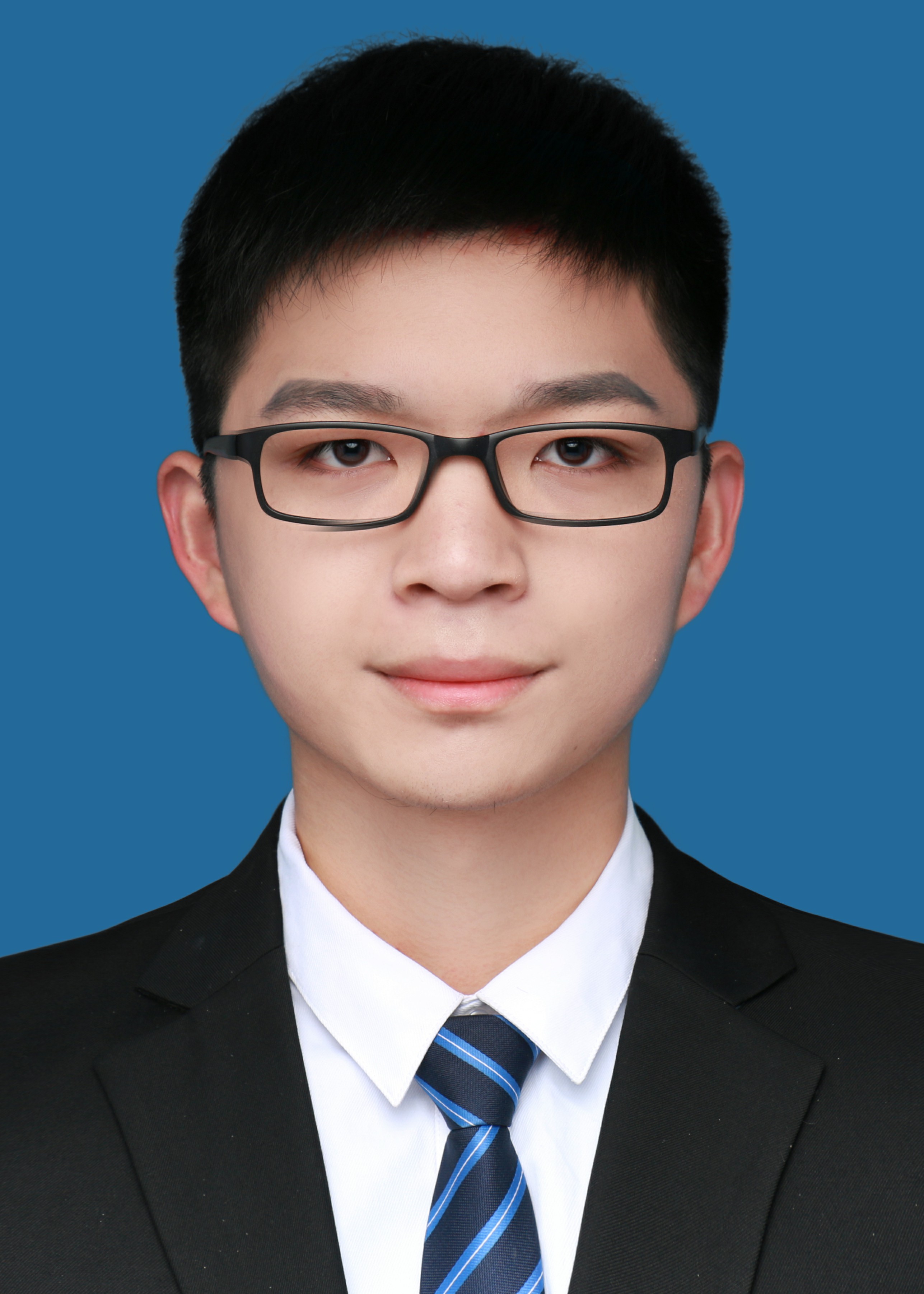}}]{Jia Li}
2016 undergraduate student, School of
Computer and Information, Hefei University of Technology.The main
research direction is natural language processing and emotional
dialogue generation.
\end{IEEEbiography}

\begin{IEEEbiography}[{\includegraphics[width=1in,height=1.25in,clip,keepaspectratio]{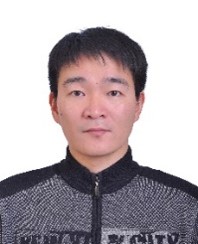}}]{Xing Wei}
Associate professor at Hefei University of Technology. His research interests include deep learning and Internet of things engineering, driverless solutions and so on.Corresponding author of this paper.
\end{IEEEbiography}

\begin{IEEEbiography}[{\includegraphics[width=1in,height=1.25in,clip,keepaspectratio]{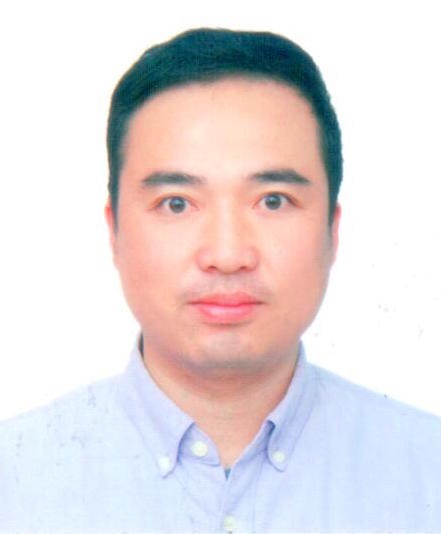}}]{Xiao Sun}
	received his double doctor's degree in Dalian University of Technology(2010) of China and the University of Tokushima(2009) of Japan. He is now working as an associate professor in AnHui Province Key Laboratory of Affective Computing and Advanced Intelligent Machine at Hefei University of Technology. His research interests include Affective Computing, Natural Language Processing, Machine Learning.
\end{IEEEbiography}

\begin{IEEEbiography}[{\includegraphics[width=1in,height=1.25in,clip,keepaspectratio]{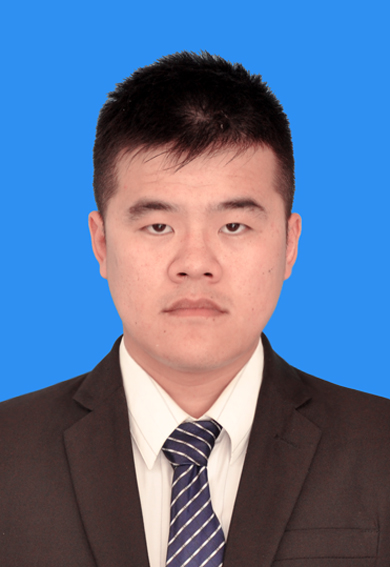}}]{Guoqiang Yang}
	Yang Guoqiang, male, master student, the main research direction is image processing and computer vision.
\end{IEEEbiography}

\begin{IEEEbiography}[{\includegraphics[width=1in,height=1.25in,clip,keepaspectratio]{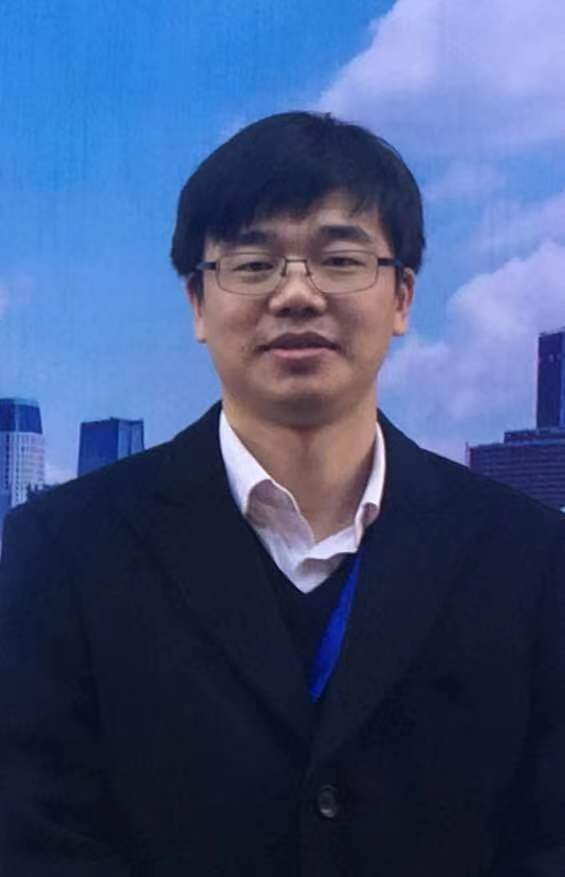}}]{Changliang Li}
Changliang Li received the Ph.D. degree from the Institute of Automation, Chinese Academy of Science, China, in 2015. Since 2018, he is currently the head with the Kingsoft institute of Artificial Intelligence. He has published widely in artificial intelligence and deep learning research. His current research interests include Deep Learning, Natural Language Processing and Data Mining.
\end{IEEEbiography}
%




\end{document}